\documentclass[11pt]{article}
\usepackage{coling2020}
\usepackage{times}
\usepackage{url}
\usepackage{latexsym}
\usepackage{amsfonts}
\usepackage{multirow}

\usepackage{graphicx}
\usepackage{float}
\usepackage{amsmath}

\colingfinalcopy % Uncomment this line for the final submission

\title{Unleashing the Power of Neural Discourse Parsers -- \\A Context and Structure Aware Approach Using Large Scale Pretraining}

\author{Grigorii Guz, Patrick Huber and Giuseppe Carenini\\
  Department of Computer Science \\
  University of British Columbia \\
  Vancouver, BC, Canada, V6T 1Z4 \\
  {\tt \{gguz, huberpat, carenini\}@cs.ubc.ca}}

\date{}

\begin{document}
\maketitle
\begin{abstract}
RST-based discourse parsing is an important NLP task with numerous downstream applications, such as summarization, machine translation and opinion mining. In this paper, we demonstrate a simple, yet highly accurate discourse parser, incorporating recent contextual language models. Our parser establishes the new state-of-the-art (SOTA) performance for predicting structure and nuclearity on two key RST datasets, RST-DT and Instr-DT. We further demonstrate that pretraining our parser on the recently available large-scale ``silver-standard" discourse treebank MEGA-DT provides even larger performance benefits, suggesting a novel and promising research direction in the field of discourse analysis. 
\end{abstract}

\section{Introduction}
\blfootnote{
    %
    % for review submission
    %
    % % final paper: en-uk version 
    %
    % \hspace{-0.65cm}  % space normally used by the marker
    % This work is licensed under a Creative Commons 
    % Attribution 4.0 International Licence.
    % Licence details:
    % \url{http://creativecommons.org/licenses/by/4.0/}.
    % 
     % final paper: en-us version 
     \hspace{-0.65cm}  % space normally used by the marker
     This work is licensed under a Creative Commons 
     Attribution 4.0 International License.
     License details:
     \url{http://creativecommons.org/licenses/by/4.0/}.
}
Discourse parsing is an important upstream task within the area of Natural Language Processing (NLP) 
which has been an active field of research over the last decades. In this work, we focus on discourse representations for the English language, where most research %on the discourse analysis of English language 
has been surrounding one of the two main theories behind discourse, the Rhetorical Structure Theory (RST) proposed by \newcite{mann1988rhetorical} or interpreting discourse according to PDTB \cite{prasadpenn}. While both theories have their strengths, the application of the RST theory, encoding documents into complete constituency discourse trees \cite{morey2018dependency}, has been shown to have many crucial implications on real world problems. A tree is defined on a set of EDUs (Elementary Discourse Units), approximately aligning with clause-like sentence fragments, acting as the leaves of the tree. Adjacent EDUs or sub-trees are hierarchically aggregated to form larger (possibly non-binary) constituents, with internal nodes containing (1) a nuclearity label, defining the importance of the subtree (rooted at the internal node) in the local context and (2) a relation label, defining the type of semantic connection between the two subtrees (e.g., Elaboration, Background). In this work, we focus on structure and nuclearity prediction, not taking relations into account. Previous research has shown that the use of RST-style discourse parsing as a system component can enhance important tasks, such as sentiment analysis, summarization and text categorization \cite{bhatia2015better,nejat2017exploring,hogenboom2015using,gerani2014abstractive,ji2017neural}. More recently, it has also been suggested that discourse structures obtained in an RST-style manner can further be complementary to learned contextual embeddings, like the popular 
BERT approach \cite{devlin2018bert}. Combining both approaches has shown to support tasks where linguistic information on complete documents
is critical, such as argumentation analysis \cite{chakrabarty2019ampersand}. 
Even though discourse parsers appear to enhance the performance on a variety of tasks, the full potential of using more linguistically inspired approaches for downstream applications has not been unleashed yet. The main open challenges of integrating discourse into more NLP downstream tasks and to deliver even greater benefits have been a combination of (1) discourse parsing being a difficult task itself, with an inherently high degree of ambiguity and uncertainty and (2) the lack of large-scale annotated datasets, rendering the initial problem more severe, as data-driven approaches cannot be applied to their full potential.

The combination of these two limitations has been one of the main reasons for the limited application of neural discourse parsing for more diverse downstream tasks. While there have been neural discourse parsers proposed \cite{braud-etal-2017-cross-lingual,yu2018transition,mabona-etal-2019-neural}, they still cannot consistently %strongly 
outperform traditional approaches when applied to the RST-DT dataset, where the amount of training data is arguably insufficient for such data-intensive approaches. 
%due to the extra effort to integrate discourse trees into models as well as two major problems, the big breakthrough in the usage of discourse parsing has still not happened. 

In this work, we alleviate the restrictions to the effective and efficient use of discourse as mentioned above by introducing a novel approach combining a newly proposed large-scale discourse treebank with our data-driven neural discourse parsing strategy.
More specifically, we employ the novel MEGA-DT ``silver-standard" discourse treebank published by \newcite{huber2020MEGA} containing over 250,000 discourse annotated documents from the Yelp'13 sentiment dataset \cite{tang2015document}, nearly three orders of magnitude larger than commonly used RST-style annotated discourse treebanks (RST-DT \cite{carlson2002rst}, Instructional-DT \cite{subba2009effective}). Given this new dataset with a previously unseen number of full RST-style discourse trees, we revisit the task of neural discourse parsing, which has been previously attempted by  \newcite{yu2018transition} and others with rather limited success. We believe that one reason why previous neural models could not yet consistently outperform more traditional approaches, heavily relying on feature engineering \cite{wang-etal-2017-two}, is the lack of generalisation when using deep learning approaches on the small RST-DT dataset, containing only 385 discourse annotated documents. This makes us believe that using a more advanced neural discourse parser in combination with a large training dataset can lead to significant performance gains. %, but also across datasets, capturing more general discourse phenomena and avoiding potential overfitting on the training corpus.
Admittedly, even though MEGA-DT contains a huge number of datapoints to train on, it has been automatically annotated, potentially introducing noise and biases, which can negatively influence the performance of our newly proposed neural discourse parser when solely trained on this dataset. A natural and intuitive approach to make use of the neural discourse parser and both datasets (``silver-standard" and gold-standard) is to combine them during training, pretraining on the large-scale ``silver-standard" corpus and subsequently fine-tuning on RST-DT or further human annotated datasets. This way, general discourse structures could be learned from the large-scale treebank and then enhanced with human-annotated trees. With the results shown in this paper strongly suggesting that our new discourse parser can encode discourse more effectively, we hope that our efforts will prompt researchers to develop more linguistically inspired applications based on our discourse parser. % for downstream models in the area of NLP. 
Our contributions in this paper are:
%\begin{itemize}
    %\item 
    
\textbf{(1)} We propose a novel neural discourse parsing architecture which combines multiple lines of previous work in a single framework.
    %\item 
    %\item
    
\textbf{(2)} We combine %the two treebanks 
a large-scale ``silver-standard" treebank (MEGA-DT) with small, domain-specific gold-standard treebanks in a neural way during the training process, by initially pretraining on the large (domain-independent) dataset and subsequently fine-tuning on the dataset within the domain itself. 

\textbf{(3)} We apply the neural discourse parser on %the large-scale ``silver-standard" discourse corpus as well as small-scale gold-standard treebanks
two commonly used disocurse treebanks (RST-DT and Instruction-DT), showing large performance improvements of our model over previous state-of-the-art approaches.% to compare the performance individually for both datasets.

%\end{itemize}
%on how to train a neural discourse parser with large scale ``silver-standard" discourse trees. With this new approach, we (1) drastically increase the amount of available training data available for discourse parsers is not sufficiently large to train modern, data-driven deep learning approaches for the task, hindering the application of new methodologies and (2) the shift in domain between the discourse parsers training data and the domain of application deminishes the applicability and performance of generated discourse trees for any domain outside of news (RST-DT), instructions (Instructional-DT) and a few other domains.
\section{Related Work}
\label{rel_work}
The field of discourse parsing has been mainly dominated by traditional machine learning approaches, frequently outperforming initial attempts to apply deep learning and neural networks to the task. Independent of the specific approach used, three  general methodologies have been followed to learn discourse trees from small datasets, such as RST-DT \cite{carlson2002rst} or Instructional-DT \cite{subba2009effective}: (1) Top-down discourse parsers, splitting the document into non-overlapping text-constituents starting from the representation of the complete discourse down to individual EDUs, assigning the two resulting sub-spans a nuclearity attribute and predicting the relation holding between the sub-trees \cite{lin2019unified}. (2) Bottom-up parsing, starting from the discourse-segmented list of EDUs and aggregating two adjacent units in every step. This approach is mostly realized using the CKY dynamic programming strategy to obtain optimal trees as in \newcite{joty2015codra} and \newcite{li-etal-2016-discourse} or using a greedy method \cite{hernault2010hilda}. (3) A frequently used and more locally inspired approach of bottom-up discourse parsing using the linear shift-reduce framework, adopted from previous work in syntactic parsing. 
While the current traditional state-of-the-art discourse parser by \newcite{wang-etal-2017-two} uses the bottom-up shift-reduce method predicted by two separate Support Vector Machines (SVMs) for structure/nuclearity prediction and relation estimation, neural models \cite{yu2018transition,braud-etal-2017-cross-lingual} utilize multi-layer perceptrons (MLP) for classifying possible actions. In our work we also follow the bottom-up shift-reduce strategy, with the detailed description of our system provided in the following section.

Besides the active research area on discourse parsing, a second line of work has emerged recently, trying to generate large-scale discourse treebanks through automated annotations from downstream tasks, such as sentiment analysis \cite{huber2020MEGA}, text classification \cite{liu-lapata-2018-learning}, summarization \cite{liu2019single} and fake news detection \cite{karimi2019learning}. The majority of these approaches follows the intuition that discourse trees can be inferred from downstream tasks by predicting latent representations during the learning process of the task itself \cite{liu-lapata-2018-learning,karimi2019learning,liu2019single} in an end-to-end manner. However, recent work by \newcite{ferracane2019evaluating} has shown that the trees resulting  from this approach are not only poorly aligned with general discourse structures, but furthermore are oftentimes too shallow. A rather different strategy has been employed by \newcite{huber2020MEGA}, trying to explicitly generate a discourse augmented treebank through distant supervision from sentiment annotations in combination with Multiple-Instance Learning (MIL) and a CKY bottom-up tree generation approach. While the resulting MEGA-DT dataset has only been proposed and released recently, the authors show promising results in their work, reaching the best inter-domain performance when comparing their dataset against RST-DT and the Instruction-DT. This leads us to believe that their treebank does not only learn sentiment-related information, but can also be used to infer general discourse structures on a large scale. We will further evaluate this in section \ref{results}. 

%Traditional discourse parsers\\
%Neural discourse parsers\\
%Datasets (inferring trees, ours)\\
%Sentiment analysis\\

\section{Neural Discourse Parsing}
%With only few neural discourse parsers previously proposed and applied to the RST-DT dataset \cite{yu2018transition,mabona-etal-2019-neural,braud-etal-2017-cross-lingual}, 
We follow the well-established bottom-up shift-reduce aggregation principle, as previously shown effective for traditional discourse parsers such as \cite{ji2014representation,wang-etal-2017-two} as well as neural approaches \cite{braud-etal-2017-cross-lingual,yu2018transition}. In this section, we will first introduce the general principle of shift-reduce parsing and define the necessary data structures and actions available to our system. Based on the general description, we will subsequently describe the approach taken to execute a single step in the linear-time model. 
%\begin{figure*}
%  \centering{\includegraphics[width=1\linewidth]
%  {imgs/all_actions_cls.png}}
%    \caption{(a) Example of a Shift action - top element of the Queue gets moved to the top of the Stack. (b) Example of a Reduce action - top 2 elements of a stack are assembled into a subtree, with the top stack element becoming the right subtree. (c) Example of input to RoBERTa corresponding to actions in (a) and (b). Individual EDUs participating in the current action are colour-coded. Note that since EDUs 1 and 2 form a subtree, their spans are concatenated, and that $S_1$ span is padded with [MASK] tokens at the end to match its desired length of 240 tokens. 
%    }
%    \label{all-actions}
%\end{figure*}

%\begin{figure}
%  \centering{\includegraphics[width=1\linewidth]
%  {imgs/sr_p.png}}
%    \caption{Alternative? Showing that we concatenate the EDU text rather than creating a dense representation or trees on the stack}
%    \label{all-actions}
%\end{figure}

%\begin{figure}
%  \centering{\includegraphics[scale=0.18]
%  {imgs/shift_action.png}}
%    \caption{Example of a shift action in SR architecture.
%    }
%    \label{dep-tree-allsteps}
%\end{figure}
%\begin{figure}
%  \centering{\includegraphics[scale=0.18]
%  {imgs/reduce_action.png}}
%    \caption{Example of a reduce action in SR architecture.
%    }
%    \label{dep-tree-allsteps}
%\end{figure}
%\subsection{Basic architecture}
\subsection{General Shift-Reduce Architecture}
\label{architecture}
\begin{figure}
  \centering{\includegraphics[width=1\linewidth]
  {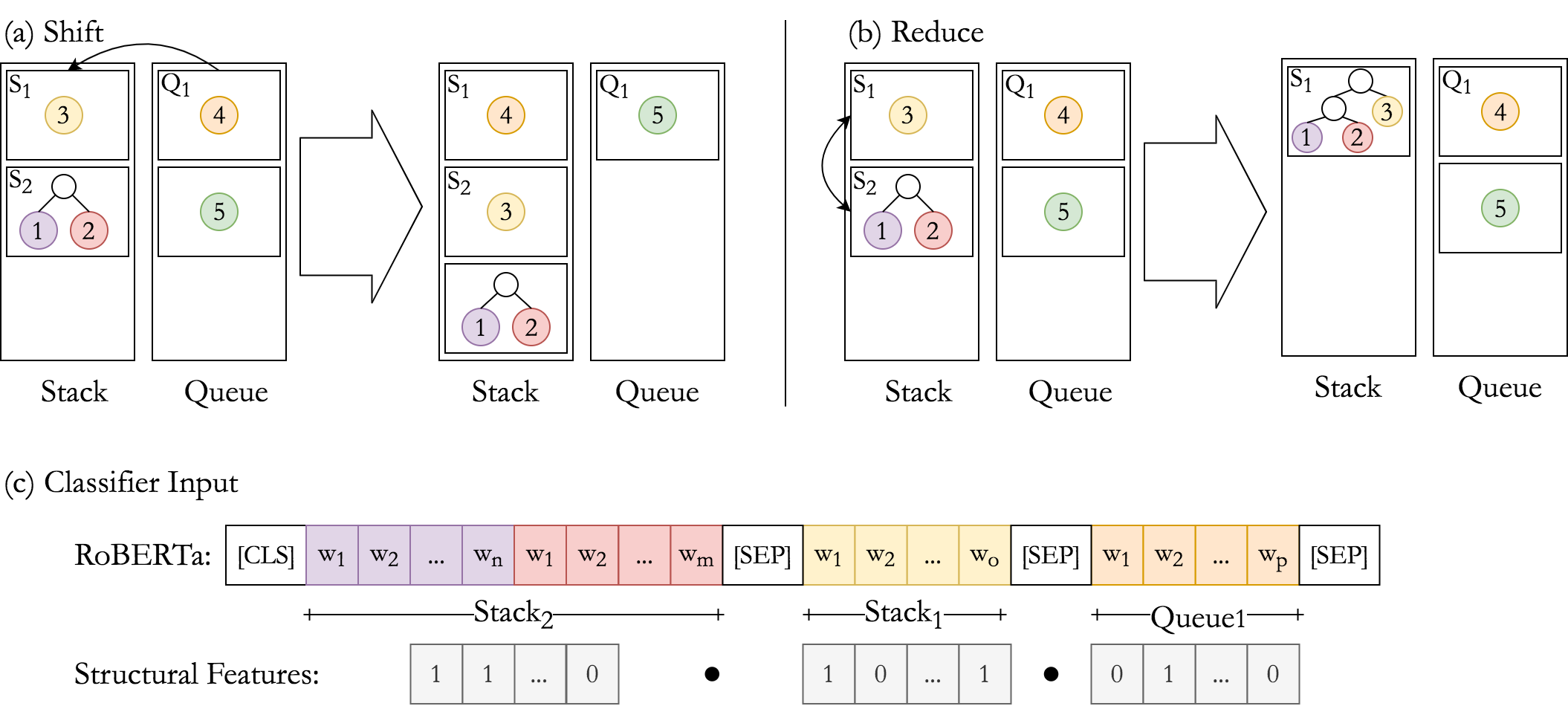}}
    \caption{(a) Example Shift action --- Top element of the Queue gets moved to the top of the Stack. (b) Example Reduce action --- Top 2 elements of a stack are assembled into a subtree%, with the top stack element becoming the right subtree
    . (c) Example of input to our classifier, consisting of the RoBERTa string-encoding and numerical, structural features. Note that since EDUs 1 and 2 form a subtree, their spans are concatenated.}
    \label{all-actions}
\end{figure}

%We use the standard transition-based shift-reduce parsing architecture \cite{marcu1999decision} as the base algorithm of our model. 
The transition-based shift-reduce parsing architecture traditionally consists of two data structures (a queue and a stack), interacting through a set of possible actions 
categorized into shift and reduce actions. This architecture is illustrated in Figure \ref{all-actions}(a, b).

\paragraph{The Queue}initially contains the EDUs of the complete document in the natural, sequential order, obtained either from manual annotation or off-the-shelf discourse segmenters (e.g.  \cite{li-sun-joty-ijcai-18}). Depending on the action performed by the parser, the top element on the queue is either read or moved to the stack. At the end of the parsing process, the queue must be empty.

\paragraph{The Stack}represents the previously processed part of the document (also in natural order). At the beginning of the process, the stack is empty and is subsequently filled and aggregated according to the system's actions. After the parsing process is completed, the stack contains the complete, aggregated document as a single discourse tree. 
 
%\begin{itemize}
%    \item A \textit{Queue} that contains all EDUs within a document, ordered from the first to the last one.
%    \item A \textit{Stack} that takes in the EDUs from the queue and keeps track of the document RST subtrees built so far.
%\end{itemize}
%To build the document tree, we use the following Stack-Queue operations:
%\begin{itemize}
%    \item SHIFT operation places the top EDU from the Queue on the top of the Stack
%    \item REDUCE-X, where $X\in\{\text{NN, NS, SN}\}$ merges the top two subtrees from the Stack into a single subtree with the corresponding nuclearity type, and places that subtree back on top of the Stack.
%\end{itemize}

\paragraph{The Shift operation} delays aggregations of sub-trees at the beginning of the document by popping the top input node (EDU) off the queue and pushing it onto the stack. %This action is only valid as long as the queue is non-empty. 
The shift-reduce algorithm needs to set hard constraints to only allow shift operations if the queue still contains unprocessed nodes.

\paragraph{The Reduce-X operation}is used to aggregate the top two partial trees ($S_1, S_2$) on the stack into a single representation ($S_{1,2}$). For complete RST-style discourse trees, each reduce action needs to further define a nuclearity assignment $X_N\in\{\text{NN, NS, SN}\}$ to the sub-tree covering $S_{1,2}$ and a relation $X_R\in\{\text{Elaboration, Contrast, ...}\}$ holding between them. %The stack must contain at least two elements for this action to be executable.
In this work, we limit the scope of the reduce action to solely predict the nuclearity assigment $X_N$, as the MEGA-DT treebank currently only provides partial discourse trees, not incorporating the relation assignment.

%Given a sequence of shift and reduce actions on the stack and queue of a document, a corresponding tree can be generated and vice versa. 
While the specific implementation of the stack and queue components are mostly fixed, the selection of shift- and reduce-actions can be realized with rule-based approaches \cite{marcu-discourse} or a variety of machine learning models, such as Support Vector Machines (SVMs) \cite{ji2014representation} or neural classifiers \cite{yu2018transition}.
The shift-reduce action selection classifier used in this work is explained below. %in section \ref{operation_explanation}.
%responsible for selecting an appropriate action at each parsing step operates as follows.
 
%\subsection{Action classifier parametrization}
\subsection{Shift-Reduce Action Classifier}
\label{operation_explanation}
%As mentioned in section \ref{architecture}, 
%The action classifier %classifier to predict the next action to generate the tree-representation 
%can be implemented in manifold ways (e.g. SVMs \cite{ji2014representation}, LSTMs \cite{yu2018transition} or Tree-LSTMs \cite{mabona-etal-2019-neural}).
%with successful approaches in previous work applying SVMs \cite{ji2014representation}, LSTMs \cite{yu2018transition} and Tree-LSTMs \cite{mabona-etal-2019-neural}. %taking partial information from the stack and queue into account when making a decision. %While LSTM networks have shown good performance in capturing sequential information in many domains and for many applications \cite{10.1145/1143844.1143891}, 
%However, BERT \cite{devlin2018bert} has recently been shown to capture more contextually inspired information and has consistently outperformed previous state-of-the-art systems \cite{zhong-etal-2019-searching,zhou-zhao-2019-head}. 
In this work, we take advantage of recent success of the BERT-inspired models on the task of language modeling, employing the distilled version \cite{sanh2019distilbert} of the large-scale RoBERTa \cite{liu2019roberta} language model as the base for our action classifier. At each step in the shift-reduce framework, we predict the next action by encoding the top two elements of the Stack ($S_1$ and $S_2$) and the top element of the Queue ($Q_1$) \cite{wang-etal-2017-two} using the RoBERTa model as well as additional structural features. 

\paragraph{RoBERTa-based semantic features:} To align with the input format requirements of the RoBERTa model, we first encode the joint representation of the three components under consideration (namely $S_1$, $S_2$ and $Q_1$) into a single string
\begin{center}
    $s = [CLS] \| S_2 \| [SEP] \| S_1 \| [SEP] \| Q_1 \| [SEP]$\\
\end{center}
with $\|$ denoting the concatenation and following the standard RoBERTa syntax, $[CLS]$ and $[SEP]$ representing the sequence-classification and end-of-sequence tokens respectively. Please note that $S_1$, $S_2$ and $Q_1$ are not the typically used dense representations of sub-trees or EDUs, but solely contain the textual representation of a sub-tree or EDU as a flat sequence of words.
%following the standard RoBERTa syntax shown in Figure \ref{fig:roberta-input}. 
%The start of the sequence is thereby described by a $[CLS]$-token, followed by the first input (the textual representation of S\textsubscript{2}). After the complete EDU representation of S\textsubscript{2}, a single $[SEP]$-token is used as the EDU separator. S\textsubscript{1} and Q\textsubscript{1} are appended in a similar fashion.

%\subsubsection{Extracting semantic features}
%We first arrange the EDU spans of the aforementioned elements as a single string of the form:
%\begin{figure*}
%  \centering{\includegraphics[width=1\linewidth]
%  {imgs/cls_example.png}}
%    \caption{Example of input to RoBERTa. 
%    }
%    \label{fig:roberta-input}
%\end{figure*}

%\begin{align}
%    s = \hspace{1mm} &\text{[CLS] S$_2$ EDUs}; \\
%        \nonumber&\text{[SEP] S$_1$ EDUs}; \\
%        \nonumber&\text{[SEP] Q$_1$ EDUs}
%\end{align}
%where ; denotes concatenation, and [SEP] and [CLS] are respectively end-of-sequence and sequence classification tokens (see Figure \ref{roberta-input} for example input).  
Since the input sequence-length of the RoBERTa language model is by default bound by 512 tokens and the elements on the stack ($S_1$, $S_2$) represent increasingly large sub-trees (and therefore text-spans) with every additional reduce operation executed, we restrict the length of $S_1$ and $S_2$ to a maximum of 240 words each. Specifically, if one of the constituents exceeds 240 tokens, the concatenation of the 120 leading and trailing words is chosen as the span's representation. The decision to retain the leading and trailing parts of a span %when exceeding the maximum number of tokens 
comes from the observation that those parts often contain important cue words which signal explicit discourse relations \cite{prasadpenn}\footnote{As opposed to implicit discourse relations, which can only be inferred from the complete semantics of spans.}. Furthermore, if the length of $S_1$ or $S_2$ is below the maximum number of 240 tokens, it is padded with [MASK] tokens %to match the desired length of 240
 to preserve absolute positions in the RoBERTa input, which is important since RoBERTa uses absolute positional embeddings. 
%The intuition is that the spans' beginnings and ends often contain cue words that indicate the constituency with adjacent spans. 
The top element on the queue (Q$_1$) is truncated or extended to the leftover capacity of 28 tokens in the same way, which clearly suffices for Queue-elements, only containing single EDUs which are 13 words on average for the RST-DT corpus and 7 words for MEGA-DT. Hence, the remaining space of 480 tokens is split equally among the stack elements, as described above. The embedding $c$ for the current stack and queue configuration is computed as follows:
%a sequence of token embeddings $v \in \mathbb{R}^{n \times 768}$ from the last layer of RoBERTa:
\begin{align}
   v = \operatorname{RoBERTa}(s)
\end{align}
with $c = v[0] \in \mathbb{R}^{768}$ encoding the full span representation at the index of the [CLS]-token.
%and the embedding for the [CLS] token $c = v[0] \in \mathbb{R}^{768}$ is utilized as the full span representation. 

%\subsubsection{Extracting text structure features}

\paragraph{Structural features:} %Utilizing the powerful RoBERTa language model to encode the joint representation of the top stack- and queue-elements allows us to take advantage of the contextual nature of BERT models. However, 
Previous successful approaches to traditional discourse parsing \cite{joty2015codra,ji2014representation} have shown that the structural organization of a document into sentences and paragraphs plays a crucial role when predicting discourse, with \newcite{joty2015codra} giving strong intuition for their usefulness by showing that less than $5\%$ of discourse subtrees violate sentence boundaries in the RST-DT corpus. To explicitly model these structural features, we use the same organizational features as \newcite{wang-etal-2017-two} 
%Similarly to \newcite{wang-etal-2017-two}, we extract text structure features in order 
to determine where a span is positioned within the document as well as relative to adjacent constituents. More specifically, for all three spans $S_1$, $S_2$ and $Q_1$, we extract two sets of features: \textbf{(1)} Whether the span is at the beginning/end of a sentence/paragraph/document and \textbf{(2)} For each pair of adjacent spans ($(S_2, S_1)$ and $(S_1, Q_1)$) we compute the features indicating whether the pair is within a single sentence/paragraph. 
For the three constituents under consideration, this results in an ordered sequence $o$ of $28$ values. Adding a distinct embedding layer with 10 neurons $u_i = emb(o_i)$ on top of each value $o_i$ in the sequence results in a concatenated dense representation $u = u_1, ..., u_{28} \in \mathbb{R}^{280}$ for the structural features.
%Since all those features are binary, we use a pair of learnable 10-dimensional embeddings for representing feature. 
Whenever a feature cannot be computed (for example when the Queue is empty or the Stack contains a single element), we represent it as a zero-vector.
%Additionally, for each pair of adjacent spans (S$_2$+S$_1$ and S$_1$+Q$_1$) we compute the features indicating whether the span pair is within a single sentence/paragraph. Since all those features are binary, we use a pair of learnable 10-dimentional embeddings for representing feature. When a feature cannot be computed (for example when Queue is empty) we represent it as a dummy vector of zeros. The resulting features are concatenated into a single vector $s$.
%\subsubsection{Making predictions}

\paragraph{Action classification:} To predict the next shift-reduce action during the tree-generation process, the semantic  and structural features described above are concatenated (see Figure \ref{all-actions}(c)) and fed into a two-layer MLP with an intermediate GeLU \cite{hendrycks2016gaussian} activation and a final softmax layer (see eq. \ref{pred}, \ref{pred_2})  for each of the four possible actions (\textit{Shift, Reduce\textsubscript{NN}, Reduce\textsubscript{NS}, Reduce\textsubscript{SN}})
\begin{align}
    \label{pred}
    %l &= W_2(\operatorname{GeLU}(W_1(c \:\|\: u) + b_1) + b_2\\
    l &= \operatorname{MLP}(c \:\|\: u)\\
    \label{pred_2}
    p &= \operatorname{SoftMax}(l)
\end{align}
%with W\textsubscript{1} $\in \mathbb{R}^{n \times m}$, b\textsubscript{1} $\in \mathbb{R}^{n}$ and W\textsubscript{2}, b\textsubscript{2} of sizes $\mathbb{R}^{m \times 4}$ and $\mathbb{R}^{4}$ respectively. 

%The final semantic and organizational features are concatenated and fed into an MLP, which outputs the unnormalized scores $a \in \mathbb{R}^4$ for each of the four possible actions:
%\begin{align}
%    a &= W_2(GeLU(W_1[c; s]) + b_1) + b_2 \label{eq-pred} \\
%    p &= SoftMax(a)
%\end{align}
%where GeLU \cite{gelu} is the activation function used in RoBERTa.

\subsection{Neural Shift-Reduce Training Procedure} 
\label{training_section}
%Using the classification module described in the previous section we do not introduce any temporal dependencies between subsequent shift-reduce actions. This allows us to train the overall model not only after a complete discourse tree has been aggregated, but on a more fine-grained level, predicting, evaluating and back-propagating the loss of every action within the sequence and forwarding gold-label actions through the tree-generation process.Hence, 
We minimize the weighted cross-entropy loss in every parsing step individually, allowing for \textbf{(1)} more fine-grained optimization and \textbf{(2)} more parallelizable training. 

The training loss is thereby computed between the unnormalized prediction $l$ (see eq. \ref{pred}) and the respective gold-label $y \in \{\textit{Shift}, \textit{Reduce}_{NN}, \textit{Reduce}_{NS}, \textit{Reduce}_{SN}\}$. %as defined in equation \ref{loss} below.
%\begin{align}
%    \label{loss}
%    loss(l, y) = w_{y}(-l_y +  \log [\sum_j \exp(l_j)])
%\end{align}
%where $y \in \{1,2,3,4\}$ is the correct class, $l$ is as in eq. \ref{pred}, and $w_y$ denotes the weight for class $y$. 
Since there is only a single shift but three reduce actions, we weight the four output classes by factors $[\frac{3}{6}, \frac{1}{6}, \frac{1}{6}, \frac{1}{6}]$ to equally penalize an incorrect shift/reduce action.
%We maximized the weighted cross-entropy loss for predicting the correct action at each parsing step:
%\begin{align}
%    loss(l, y) = w_{y}(-l_y +  \log [\sum_j \exp(l_j)])
%\end{align}
%where $y \in \{1,2,3,4\}$ is the correct class, $l$ is as in eq. \ref{pred}, and $w_y$ denotes the weight for class $y$. Since there is a single Shift and three Reduce actions, we chose the weights to be $[\frac{3}{6}, \frac{1}{6}, \frac{1}{6}, \frac{1}{6}]$ to equally penalize an incorrect Shift/Reduce action.
%\subsection{Tree Structure Inference}
At test time, the document tree structure is constructed greedily by selecting the action with the highest probability (see eq. \ref{pred_2}) at each parsing step.

\section{Experiments}
\label{experiments}
%In this section, we evaluate our newly proposed neural shift-reduce model topology. 
We first introduce the three datasets we used to train and evaluate our model against strong baselines (section \ref{datasets}). %A special focus is put on the relatively new MEGA-DT treebank, which has been proposed recently (section \ref{datasets}). 
Further, we describe how to effectively combine diverse treebanks in a neural manner using pretraining and fine-tuning, now possible with our neural discourse parser (section \ref{comb_treebanks}). Our model hyperparameters and the respective search spaces on the development set are  presented in section \ref{hyper}, followed by the baseline models in section \ref{baselines}. The experiments section will then introduce the metrics used in this paper (section \ref{metrics}), discuss insights gathered in preliminary evaluations (section \ref{prelim}) and finally present results aggregated in section \ref{results}. 

\subsection{Datasets}
\label{datasets}
This work relies on three distinct RST-style discourse treebanks for the English language. % described below.

\paragraph{RST-DT} is the largest human-annotated RST-style discourse parsing corpus \cite{carlson2002rst}, consisting of news articles from the Wall Street Journal. The treebank contains 347 documents in the training- and 38 in the test-set. 
We further split the training portion into 90\% training data %($RST$-$DT_{train}$) 
and 10\% development data %($RST$-$DT_{test}$) 
to perform hyper-parameter and architecture optimization.

\paragraph{Instructional  Dataset} is the second human-annotated RST-style corpus used in this work, containing 176 documents. The Instructional Dataset (short: Instr-DT) is used to train and evaluate discourse parsers in the home-repair instructions domain \cite{subba2009effective}. During preprocessing, we combine multi-rooted documents using a sequence of right-branching decisions with an N-N nuclearity assignment. We randomly separate the data into a 90\% training- %($Instr$-$DT_{train}$) 
and a 10\% test-portion. %($Instr$-$DT_{test})$
Please note that our training/test split is consistent across all models except the CODRA model described below, where we report the original results published by the authors using 10-fold validation.%, based on an unknown training/test split.

\paragraph{MEGA-DT} is the first successful automatically generated ``silver-standard" discourse treebank obtained by applying distant supervision on the large-scale Yelp'13 sentiment dataset \cite{tang2015document}. Recently %proposed and 
published by \newcite{huber2020MEGA}, the treebank contains $\approx$250,000 documents with full RST-style discourse trees encompassing structure and nuclearity attributes. %While there have been manifold approaches trying to infer general discourse structures from distant supervision, 
The MEGA-DT corpus has been shown to achieve superior performance when compared to human-annotated datasets (including RST-DT) on the discourse domain-transfer task. Due to computational limitations we pretrain our model on a 52.5k subset of MEGA-DT, with 50k trees used for training and 2.5k datapoints left out for the development set.

\subsection{Combining Treebanks}
\label{comb_treebanks}
%While previous work focused on one or more of the aforementioned treebanks to train and evaluate discourse parsers, the general nature of neural networks allows us to combine datasets in a structural way, by pretraining on one corpus and fine-tuning on another. %This makes especially sense with the newly proposed, large-scale MEGA-DT dataset. %We therefore do not only show our results when training and evaluating on a single treebank, but combine multiple corpora to improve the performance further.
In a first stage of our experiments, we will verify the effectiveness of our proposed architecture by training and evaluating the parser on individual datasets. Additionally, we will pretrain the parser on MEGA-DT until it converges on the development portion and then fine-tune on RST-DT and Instr-DT to evaluate the usefulness of pretraining on silver-standard discourse trees.
%\paragraph{Intra-Domain evaluation: } The first stage of our experiments involves verifying that our simple proposed architecture is indeed effective - we will train and evaluate our parser on the same domain. Additionally, we will pretrain our parser on MEGA-DT and fine-tune on RST-DT and INSTR-DT to evaluate the usefulness of pretraining on silver-standard discourse trees.

\subsection{Hyperparameters and Training Setup} 
\label{hyper}
The hyperparameters in our model are heavily influenced by previous findings. For the RoBERTa model \cite{liu2019roberta}, %which plays a crucial role in our classification prediction, 
we use the pre-trained distilled version proposed in \newcite{sanh2019distilbert} with 6 layers containing 12 attention heads and a hidden size of 768, as implemented by \newcite{Wolf2019HuggingFacesTS}.
The structural features used as inputs for the classification module are encoded as 10-dimensional embeddings for each of the 28 organizational features. During training we use the AdamW optimizer \cite{loshchilov2018decoupled} with a learning rate of 0.001 and a weight decay value of 0.01 for both pretraining and fine-tuning. We further apply gradient norm clipping at 0.2 \cite{pascanu2013difficulty}. The learning rate was scheduled as in \newcite{NIPS2017_7181}, using 4000 warm-up steps. Due to the variable size trees in the training data, %containing different number of shift- and reduce-actions, 
we aggregate documents with identical number of EDUs into batches of size 20 during pretraining and 5 for fine-tuning. All model configurations are trained by early stopping if the performance of neither structure nor nuclearity improves  over 3 consecutive epochs on the development dataset. Our models are trained using PyTorch \cite{NIPS2019_pytorch} on a GTX 1080 Ti GPU with 11GB of memory. 
Our code and model-checkpoints %and the MEGA-DT corpus
will be made publicly available with the publication of this paper\footnote{\url{http://www.cs.ubc.ca/cs-research/lci/research-groups/natural-language-processing/Software.html}}.

\subsection{Baselines}
\label{baselines}
To evaluate the performance of our %neural
model in the context of RST-style discourse parsing, we compare it against a variety of previously proposed, competitive baselines:
%\paragraph{Right/Left Branching Baselines:} predict a binary,  fully right- or left-branching tree for every document in the dataset.
%\paragraph{Hierarchical Right/Left Branching Baselines:} are simple baselines predicting a binary, fully right- or left-branching tree on sentence-level and combine the sentence-level trees in right- or left-branching manner for every document in the dataset.

The \textbf{DPLP} parser \cite{ji2014representation} is a traditional discourse parser utilizing an SVM-classifier within the shift-reduce framework solely based on linear projections of lexical features. The \textbf{CODRA} model \cite{joty2015codra} uses an optimal CKY-based chart parser in combination with Dynamic Conditional Random Fields (CRF), separated on sentence-level. The \textbf{gCRF} model \cite{feng-hirst-2014-linear} follows a similar approach but utilizes a greedy strategy. The \textbf{Two-Stage} parser proposed by \newcite{wang-etal-2017-two} is the current SOTA system on the RST-DT structure prediction. The model uses two separate linear SVM classifiers. %This ``horizontally" separated approach reaches the best performance for the structure and nuclearity prediction when compared to previous, traditional discourse parsers. 
We use the public codebase\footnote{\url{https://github.com/yizhongw/StageDP/}} provided by \newcite{wang-etal-2017-two} and remove the relation classification module for our experiments. \textbf{Transition-Syntax:} the parser by \newcite{yu2018transition} is a neural shift-reduce parser utilizing LSTMs to generate EDU embeddings. In addition, they apply a neural dependency parser for extracting syntactic features. %This ``horizontally" separated approach reaches the best performance for the structure and nuclearity prediction when compared to previous, traditional discourse parsers. 
\textbf{Cross-Lingual} is the neural shift-reduce approach by \newcite{braud-etal-2017-cross-lingual}, utilizing RST treebanks from multiple languages and the \textbf{Top-Down-Generative} parser by \newcite{mabona-etal-2019-neural} is a recent top-down transition-based neural generative parser employing Tree-LSTM for encoding subtrees on the stack. %This ``horizontally" separated approach reaches the best performance for the structure and nuclearity prediction when compared to previous, traditional discourse parsers. 

%Even though all baselines described above use traditional approaches to solve the task of discourse parsing, there have been neural discourse parsers proposed. However, due the fact that %We note that since 
%previous neural approaches were not able to consistently outperform traditional baselines \cite{morey-etal-2017-much,yu2018transition} and we have not been able to confirm their performance, we do not take them into account in our evaluation, but only include the traditional approaches currently holding state-of-the-art performance on RST-DT and Instr-DT.

\subsection{Metrics} 
\label{metrics}
%We follow the most consistently used metric to evaluate our neural discourse parser along with the baseline systems. %, to allow for a fair and extensive comparison. 
%The vast majority of previous top-performing discourse parsers uses the average micro precision on span and nuclearity level as the metric of choice, e.g., \cite{wang-etal-2017-two,joty2015codra}. 
As suggested in a recent literature analysis by \newcite{morey-etal-2017-much}, we use the original Parseval measure to compare the micro-average F1-scores of our model with our selected baselines. To further allow additional comparisons, we also report the results with respect to RST-Parseval, currently still more commonly used in recent literature. %The metric is %generally 
%computed as the overlap of discourse structure and nuclearity predictions and the gold structure with nuclearity. %The prediction and the gold structure are thereby traversed in parallel post-order and individual nodes of the tree are compared.
%The choice of presenting the precision metric over recall and F-score has no impact on the results in our case, as manual EDU segmentation is used (see \cite{wang-etal-2017-two,joty2015codra}). 

\begin{table*}[t!]
\centering
\begin{tabular}{|l|r r|r r|}
\hline
\multirow{2}{*}{Model} & \multicolumn{2}{c|}{Structure} & \multicolumn{2}{c|}{Nuclearity}\\
 & RST-DT & Instr-DT & RST-DT & Instr-DT\\
\hline \hline 
%Right Branching & 54.64 & 58.47 & $\times$ & $\times$\\
%Left Branching & 53.73 & 48.15 & $\times$ & $\times$\\
%Hier. Right Branching\shortcite{huber2019predicting} & 70.82 & 67.86 & $\times$ & $\times$\\
%Hier. Left Branching\shortcite{huber2019predicting} & 70.58 & 63.49 & $\times$ & $\times$\\
%Majority Class & $\times$ & $\times$ & (N-S) 61.28  & (N-N) 52.33 \\
%\hline \hline
%\multicolumn{5}{|c|}{\textbf{Intra-Domain} Evaluation}\\
%\hline
DPLP\shortcite{ji2014representation} 
& 64.10
& --- 
& 54.20  
& ---\\
gCRF\shortcite{feng-hirst-2014-linear} 
& 68.60
& --- 
& 55.90  
& ---\\
CODRA\shortcite{joty2015codra} 
& 65.10 
& ---  
& 55.50 
& --- \\
Cross-Lingual \shortcite{braud-etal-2017-cross-lingual} 
& 62.70 
& ---  
& 54.50 
& ---\\

%Li\shortcite{li-etal-2016-discourse}
%& 64.50
%& ---
%& 54.00
%& ---\\
Two-Stage\shortcite{wang-etal-2017-two} 
& 70.97
&  58.86
& 57.97
& 40.00 \\
Top-Down-Generative\shortcite{mabona-etal-2019-neural} 
& 67.10 
& --- 
& 57.40 
& --- \\

%Our parser (Feats only) & $_{\pm 0.0.08}$79.95  & $_{\pm 0.14}$77.34 & $_{\pm 0.24}$48.72 & $_{\pm 0.96}$59.52 \\

%Our parser (Roberta only) & $_{\pm 0.32}$85.30  & $_{\pm 0.78}$81.66 & $_{\pm 0.41}$71.04 & $_{\pm 0.52}$65.98 \\

Our Parser 
& $_{\pm 0.27}$72.43 & 
$_{\pm 1.12}$64.55
& $_{\pm 0.56}$61.38
& $_{\pm 1.37}$44.41 \\
Our Parser + Pretraining 
& $_{\pm 0.95}$\textbf{72.94} 
& $_{\pm 0.61}$ \textbf{65.41}
& $_{\pm 0.90}$\textbf{61.86}
& $_{\pm 1.11}$\textbf{46.59} \\
\hline\hline
Our Parser (- RoBERTa) 
& $_{\pm 0.18}$59.89  
& $_{\pm 0.30}$54.68 
& $_{\pm 0.55}$33.28 
& $_{\pm 1.83}$28.36 \\

Our Parser (- Features) 
& $_{\pm 0.72}$70.61  
& $_{\pm 1.74}$63.32 
& $_{\pm 0.65}$58.37 
& $_{\pm 1.83}$44.41 \\

Our Parser (- LM Pretraining) 
& $_{\pm 0.42}$ 65.78
& $_{\pm 2.97}$ 53.50
& $_{\pm 0.58}$ 48.93 
& $_{\pm 2.69}$ 32.82  \\

%Two-Stage\shortcite{wang-etal-2017-two} & 85.98$_{\pm 0.00}$ & 77.28$_{\pm 0.00}$ & 72.40$_{\pm 0.00}$ & 60.01$_{\pm 0.00}$\\
%Our parser & 86.22$^\dagger_{\pm 0.13}$  & 81.93$_{\pm 1.09}$ & 73.02$^\dagger_{\pm 0.50}$ & 65.09$_{\pm 1.94}$ \\
%Our parser + Pretraining & \textbf{86.46}$^\dagger_{\pm 0.48}$ & 82.71$_{\pm 0.30}$ & \textbf{73.52}$^\dagger_{\pm 0.70}$ & \textbf{66.48}$_{\pm 1.10}$ \\

%\hline \hline
%\multicolumn{5}{|c|}{\textbf{Inter-Domain} Evaluation}\\
%\hline
%Two-Stage\textsubscript{RST-DT} & $\times$ & 73.57 & $\times$ & 49.78\\
%Our parser\textsubscript{RST-DT} & $\times$ & 77.89 & $\times$ & 55.16\\
%Two-Stage\textsubscript{Instr-DT} & 74.32 & $\times$ & 44.68 & $\times$\\
%Our parser\textsubscript{Instr-DT} & 76.41 & $\times$ & 48.96 & $\times$\\
%Two-Stage\textsubscript{Yelp13-DT\shortcite{huber2019predicting}} & 76.41 & 74.14 & 35.72 & 33.35\\
%Two-Stage\textsubscript{MEGA-DT} & \textsuperscript{$\dagger$}\textbf{77.82} & \textsuperscript{$\dagger$}\textbf{75.18} & \textbf{44.88} & \textsuperscript{$\dagger$}\textbf{54.87}\\
%Our Parser\textsubscript{MEGA-DT} & 79.66 & 75.00 & 47.34 & 43.86\\
%Our Parser\textsubscript{MEGA-DT+RST-DT} & --- & 75.87 & --- & 2.80\\
%Our Parser\textsubscript{MEGA-DT+INSTR-DT} & 77.70 & --- & 50.23 & ---\\
\hline\hline
Human \shortcite{morey-etal-2017-much} & 78.7 & --- &  66.8 & ---\\
\hline
\end{tabular}
\caption{Micro-averaged F1-scores for structure and nuclearity prediction using the original Parseval measure as proposed in \newcite{morey-etal-2017-much}, %performance, 
evaluated on the RST-DT and Instr-DT corpora. Best performance per column is \textbf{bold}. (%\textsuperscript{$\dagger$} statistically significant with p-value $\leq .05$ to the best baseline established using one-sided t-test, 
subscripts on results indicate standard deviation, --- non-published values)}
\label{tab:final_parseval}
\end{table*}

\begin{table*}[t!]
\centering
\begin{tabular}{|l|r r|r r|}
\hline
\multirow{2}{*}{Model} & \multicolumn{2}{c|}{Structure} & \multicolumn{2}{c|}{Nuclearity}\\
 & RST-DT & Instr-DT & RST-DT & Instr-DT\\
\hline \hline 
%Right Branching & 54.64 & 58.47 & $\times$ & $\times$\\
%Left Branching & 53.73 & 48.15 & $\times$ & $\times$\\
%Hier. Right Branching\shortcite{huber2019predicting} & 70.82 & 67.86 & $\times$ & $\times$\\
%Hier. Left Branching\shortcite{huber2019predicting} & 70.58 & 63.49 & $\times$ & $\times$\\
%Majority Class & $\times$ & $\times$ & (N-S) 61.28  & (N-N) 52.33 \\
%\hline \hline
%\multicolumn{5}{|c|}{\textbf{Intra-Domain} Evaluation}\\
%\hline
DPLP\shortcite{ji2014representation} & 82.00 & --- & 68.20 & ---\\
gCRF\shortcite{feng-hirst-2014-linear} & 84.30 & --- & 69.40 & ---\\
CODRA\shortcite{joty2015codra} & 82.60 & \textbf{82.88}  & 68.30 & 64.13\\
%Li\shortcite{li-etal-2016-discourse} & 82.20 & ---  & 66.50 & %---\\
Cross-Lingual \shortcite{braud-etal-2017-cross-lingual} & 81.30 & ---  & 68.10 & ---\\

Transition-Syntax\shortcite{yu2018transition} & 85.50 & --- & 73.10 & ---\\
Two-Stage\shortcite{wang-etal-2017-two} & 85.98 & 79.43 & 72.40 & 62.39\\

%Our parser (Feats only) & $_{\pm 0.0.08}$79.95  & $_{\pm 0.14}$77.34 & $_{\pm 0.24}$48.72 & $_{\pm 0.96}$59.52 \\

%Our parser (Roberta only) & $_{\pm 0.32}$85.30  & $_{\pm 0.78}$81.66 & $_{\pm 0.41}$71.04 & $_{\pm 0.52}$65.98 \\

Our Parser 
& $_{\pm 0.13}$86.22  & 
$_{\pm 0.56}$82.27 
& $_{\pm 0.50}$ 73.03 
& $_{\pm 1.11}$65.82 \\
Our Parser + Pretraining 
& $_{\pm 0.48}$ \textbf{86.47} 
& $_{\pm 0.30}$82.71 
& $_{\pm 0.70}$\textbf{73.53} 
& $_{\pm 1.01}$\textbf{66.59} \\
\hline\hline
Our Parser (- RoBERTa) 
& $_{\pm 0.09}$79.95  
& $_{\pm 0.15}$77.34 
& $_{\pm 0.27}$48.72 
& $_{\pm 1.07}$59.52 \\

Our Parser (- Features) 
& $_{\pm 0.36}$85.30  
& $_{\pm 0.87}$81.66 
& $_{\pm 0.46}$71.04 
& $_{\pm 0.59}$65.98 \\

Our Parser (- LM Pretraining) 
& $_{\pm 0.21}$82.89  
& $_{\pm 1.49}$76.75 
& $_{\pm 0.24}$63.22 
& $_{\pm 0.85}$59.18 \\

%Two-Stage\shortcite{wang-etal-2017-two} & 85.98$_{\pm 0.00}$ & 77.28$_{\pm 0.00}$ & 72.40$_{\pm 0.00}$ & 60.01$_{\pm 0.00}$\\
%Our parser & 86.22$^\dagger_{\pm 0.13}$  & 81.93$_{\pm 1.09}$ & 73.02$^\dagger_{\pm 0.50}$ & 65.09$_{\pm 1.94}$ \\
%Our parser + Pretraining & \textbf{86.46}$^\dagger_{\pm 0.48}$ & 82.71$_{\pm 0.30}$ & \textbf{73.52}$^\dagger_{\pm 0.70}$ & \textbf{66.48}$_{\pm 1.10}$ \\

%\hline \hline
%\multicolumn{5}{|c|}{\textbf{Inter-Domain} Evaluation}\\
%\hline
%Two-Stage\textsubscript{RST-DT} & $\times$ & 73.57 & $\times$ & 49.78\\
%Our parser\textsubscript{RST-DT} & $\times$ & 77.89 & $\times$ & 55.16\\
%Two-Stage\textsubscript{Instr-DT} & 74.32 & $\times$ & 44.68 & $\times$\\
%Our parser\textsubscript{Instr-DT} & 76.41 & $\times$ & 48.96 & $\times$\\
%Two-Stage\textsubscript{Yelp13-DT\shortcite{huber2019predicting}} & 76.41 & 74.14 & 35.72 & 33.35\\
%Two-Stage\textsubscript{MEGA-DT} & \textsuperscript{$\dagger$}\textbf{77.82} & \textsuperscript{$\dagger$}\textbf{75.18} & \textbf{44.88} & \textsuperscript{$\dagger$}\textbf{54.87}\\
%Our Parser\textsubscript{MEGA-DT} & 79.66 & 75.00 & 47.34 & 43.86\\
%Our Parser\textsubscript{MEGA-DT+RST-DT} & --- & 75.87 & --- & 2.80\\
%Our Parser\textsubscript{MEGA-DT+INSTR-DT} & 77.70 & --- & 50.23 & ---\\
\hline\hline
Human \shortcite{morey-etal-2017-much} & 88.30 & --- &  77.30 & ---\\
\hline
\end{tabular}
\caption{Micro-averaged F1-scores for structure and nuclearity prediction using RST-Parseval, %performance, 
evaluated on the RST-DT and Instr-DT corpora. Best performance per column is \textbf{bold}. (%\textsuperscript{$\dagger$} statistically significant with p-value $\leq .05$ to the best baseline established using one-sided t-test, 
subscripts on results indicate standard deviation, --- non-published values)}
\label{tab:final_rst_parseval}
\end{table*}

\subsection{Preliminary Evaluation} 
\label{prelim}
In our preliminary experiments, we evaluate a set of modelling decisions on the held-out development set, influencing the design of our final model. We obtained three useful insights during this phase:\\
%We observed that the following modelling strategies resulted in highest performance of our model on RST-DT development set: (1) 
\textbf{(1)} Adding padding to the three classifier input strings used in the RoBERTa model, extending each of them to the maximum defined length of 240 words for the two stack elements $S_1$ and $S_2$ and padding the top element on the queue ($Q_1$) to 28 words  substantially enhanced the performance of the component. We believe this is likely to be the case because RoBERTa internally uses absolute positional embeddings.\\ \textbf{(2)} Following the intuition that an incorrect shift- and reduce-action should be penalized similarly (independent of the nuclearity label), we found that weighting the loss function as described in section \ref{training_section}
%, as opposed to using uniform weighting, appeared to 
boosts the model's performance on both, the structure and nuclearity metric. \\%, when compared against uniform weighting.
\textbf{(3)} We experimented with different ways for summarizing the outputs of RoBERTa (see Figure \ref{all-actions}(c)) into a single vector. After experiments with simple and attention-based averaging of RoBERTa outputs, we found these approaches to produce sligtly worse results compared to simply using the output vector corresponding to the [CLS] token.

%\begin{table*}[ht!]
%\centering
%\scalebox{1.05}{
%\begin{tabular}{|l|r r|r r|}
%\hline
%\multirow{2}{*}{Model} & \multicolumn{2}{c|}{Structure} & \multicolumn{2}{c|}{Nuclearity}\\
%& RST-DT & Instr-DT & RST-DT & Instr-DT\\
%\hline \hline 
%Our parser (Feats only) & $_{\pm 0.08}$79.95  & $_{\pm 0.14}$77.34 & $_{\pm 0.27}$48.72 & $_{\pm 1.06}$59.52 \\
%
%Our parser (Roberta only) & $_{\pm 0.32}$85.30  & $_{\pm 0.78}$81.66 & $_{\pm 0.46}$71.04 & $_{\pm 0.59}$65.98 \\
%
%Our parser & $_{\pm 0.13}$$^\dagger$86.22  & $_{\pm 1.09}$81.93 & $_{\pm 0.50}$$^\dagger$73.03 & $_{\pm 1.97}$65.39 \\
%
%Our parser + Pretraining & $_{\pm 0.48}$$^\dagger$\textbf{86.47} & $_{\pm 0.30}$82.71 & $_{\pm 0.70}$$^\dagger$\textbf{73.53} & $_{\pm 1.08}$$^\dagger$\textbf{66.59} \\
%
%\hline
%\end{tabular}}
%\caption{Ablation study using average micro precision results for structure- and nuclearity-prediction, %performance, 
%evaluated on the RST-DT and Instr-DT corpora. Best performance per column is \textbf{bold}. Subscripts on results indicate standard-deviation)}
%\label{tab:ablation}
%\end{table*}

\subsection{Results} 
\label{results}
The final results of our evaluation are presented in Tables \ref{tab:final_parseval} and \ref{tab:final_rst_parseval}. The two tables contain slightly different subsets of competitive discourse parsers from previous work, depending on the metric on which the original authors evaluate their models.  %with respect to only one of the aforementioned metrics. 
The reported scores %for previous work 
were either taken from the original paper or the literature survey by \newcite{morey-etal-2017-much}\footnote{Even though the authors of the Two-Stage parser only report RST-Parseval scores on RST-DT, we also evaluate their approach on Instr-DT and with respect to the original Parseval metric.}.  %The evaluation is thereby separated into three columns. 

The left-most column in the first sub-table contains the models in the comparison, %, with the simple hierarchical left- and right-branching baselines on top, along 
showing previously proposed baselines along with two versions of our new neural discourse parser, with and without pretraining. The center column contains the evaluation results on the structure prediction task for the two test datasets (RST-DT and Instr-DT). The right-most column shows the performance for each of the models on the nuclearity prediction task, again subdivided for the two evaluation datasets. The %first row block in the each table contains the previous parsers and our final models, while the 
second sub-table contains the results for various ablations of our model and the bottom sub-table shows human results on the tasks. For all of our models we report the average performance as well as the standard deviation on each metric over five independent runs. %and therefore marks the upper-bound for automated approaches. 
We compare the two versions of our neural discourse parser against the best-performing, previously proposed model for each of the four prediction tasks, %In our case, this means 
meaning we compare the performance of our model, for example on the RST-DT dataset, against the current SOTA model on RST-DT structure-prediction by \newcite{wang-etal-2017-two} and for nuclearity against \newcite{yu2018transition}. The best performing baseline on the Instr-DT dataset is the CODRA model \cite{joty2015codra}. Please note again the different evaluation procedure on Instr-DT that was used for CODRA model \cite{joty2015codra}. For a more direct comparison, please see our Instr-DT results against the Two-Stage parser \cite{wang-etal-2017-two}, which utilizes the same split as ours.  %, which will be used to compare our model against on this dataset. %Please note that as already mentioned earlier, we do not compare our model against the previously proposed neural model by \newcite{yu2018transition} %due to the inconsistent performance across structure and nuclearity and furthermore 
%because we could not confirm the performance published in their paper in our preliminary studies.

When examining our final evaluations shown in Tables \ref{tab:final_parseval} and \ref{tab:final_rst_parseval} it becomes clear that our newly proposed neural discourse parser reaches the highest performance on all measures except the structure prediction on the Instr-DT dataset. 
%While this already shows the superior performance of our model, we further employ a significance analysis on the measures to strengthen our results. To this end, we run all models with publicly available implementations for the training- and evaluation-procedure, as well as our models with 5 randomly initialized seeds and compute the statistical significance, using the one-sided t-test between our results and the best baseline. For the model with additional pretraining we use a single pretraining run and fine-tune the model on the target domain with 5 independent random initialization. 
We observe that our model strongly outperforms the SOTA approach on the RST-DT structure prediction by \newcite{wang-etal-2017-two}. Furthermore, pretraining on the MEGA-DT treebank leads to further improvement with respect to the mean scores over independent runs. %, however, the improvement is not significant over the performance of our approach without pretraining on MEGA-DT.

On the Instr-DT dataset, our parser achieves a result similar to the model of \newcite{joty2015codra} on the structure prediction task and substantially outperforms the SOTA baseline on the nuclearity measure when pretraining is applied. 
%As we are not able to perform statistical significance testing against the CODRA discourse parser, due to no publicly available implementation of the training procedure, we resort to showing the standard-deviation of our system as an indication of the stability of our performance (shown as subscript of the results of our system).  
Nonetheless, a particularly important result is that our system produces consistently strong performance across multiple domains, which neither of the top-performing traditional systems \cite{wang-etal-2017-two,joty2015codra} managed to demonstrate. This serves as an indication that employing large-scale language models alleviates the need for extensive manual feature engineering employed by these systems for RST discourse parsing.

In addition, we perform an ablation study of our system to analyze the importance of each individual component of our parser. The first row in the second sub-tables illustrates the results when only organizational features are used, while the second row shows the impact of removing the features and only using RoBERTa for the action classification. Finally, the third row contains the performance of our system with organizational features and a randomly initialized RoBERTa model component. 

We obverse that removing the organizational features results in a noticeable drop in performance, implying the importance of encoding the document structure explicitly. Unsurprisingly, removing the RoBERTa feature extractor leads to a large performance drop, far below the competitive baselines, since this version of our system does not take discourse connective words into account. Finally, we demonstrate the importance of LM pretraining and pretrained word embeddings in the last row of the ablation sub-table. While this system performs on par with traditional systems in respect to structure prediction, most likely because of the organizational features, it demonstrates inferior performance on the nuclearity prediction task, which (even in the easiest scenarios) requires knowledge of more high-level concepts, such as sentence coordination and subordination. In more advanced cases, it plausibly requires knowledge about the author's communicative goal. The overall difficulty of this task is reflected in the relatively low human evaluation scores shown in the last row. Our results can be summarized as follows:\\
\textbf{(1)} Our proposed approach %is the first to achieve 
achieves state of the art performance on both the RST-DT and the Instr-DT datasets. \textbf{(2)} Applying large-scale language models leads to stronger results and higher domain adaptivity in RST discourse parsing. \textbf{(3)} Pretraining the discourse parser on the large-scale ``silver-standard" MEGA-DT treebank enhances the performance and supports the ability of the neural parser to generalize across multiple datasets and domains.\\

\section{Conclusions} 
In this work, we proposed a rather simple, yet highly effective discourse parser, utilizing recent neural BERT-based language models in combination with structural features. The integration of those input-features within a standard shift-reduce framework as well as an unprecedented use of recent large-scale ``silver-standard" discourse parsing datasets for pretraining reaches a new state-of-the-art performance on both, the RST-DT and Instr-DT treebanks. We show that our new, neural discourse parser already achieves better or similar performance when trained and evaluated on the RST-DT and Instr-DT datasets, however, the consistent and significant SOTA result is reached when incorporating pretraining on the MEGA-DT corpus. %This refutes the previous findings of \newcite{morey-etal-2017-much}, stating that neural techniques such as word embeddings provide little to no gains for this task. %We further demonstrated the gains achieved by 
The presented pretraining %even on the small subset of
approach on the silver-standard MEGA-DT dataset %, which 
also further validates the usefulness of additional supervision for this task and calls for more work in that area. 

\section{Future Work}

As directions for future work, we plan to run experiments with larger language models, as our lightweight RoBERTa model only contains 82M parameters, while top-performing language models such as BERT-Large utilize an order of magnitude more parameters. %This direction has a promising potential for better generalization, but requires significant computational resources. 
We also want to explore neural parsing strategies besides the shift-reduce framework trained on the large-scale MEGA-DT treebank, comparable to the recently proposed top-down neural discourse parsing architecture by \newcite{mabona-etal-2019-neural}. Further, we plan to extend our framework to also predict relations, generating complete discourse trees. Another line of future work is to evaluate the effectiveness of pretraining on other, more shallow discourse analysis frameworks and datasets such as PDTB, only containing flat discourse trees, which might potentially require new approaches for silver-standard annotation. Lastly, in the short term we plan to overcome our computational limitations and pretrain our model on the full MEGA-DT. After that, we also want to generally venture into even larger pretrained treebanks generated according to \newcite{huber2020MEGA}, taking into account more diverse sentiment datasets (such as IMDB \cite{diao2014jointly} and Amazon reviews \cite{zhang2015character}) to extend the size and generality of the pretraining approach and eventually enhance the overall performance of our parser.

\section*{Acknowledgments}
We thank the anonymous reviewers and the UBC-NLP group for their insightful comments and suggestions. 
This research was partially supported by the Language \& Speech Innovation Lab of Cloud BU, Huawei Technologies Co., Ltd.

% include your own bib file like this:
\bibliographystyle{coling}
\bibliography{coling2020}

\begin{thebibliography}{}

\bibitem[\protect\citename{Bhatia \bgroup et al.\egroup
  }2015]{bhatia2015better}
Parminder Bhatia, Yangfeng Ji, and Jacob Eisenstein.
\newblock 2015.
\newblock Better document-level sentiment analysis from {RST} discourse
  parsing.
\newblock In {\em Proceedings of the 2015 Conference on Empirical Methods in
  Natural Language Processing}, pages 2212--2218.

\bibitem[\protect\citename{Braud \bgroup et al.\egroup
  }2017]{braud-etal-2017-cross-lingual}
Chlo{\'e} Braud, Maximin Coavoux, and Anders S{\o}gaard.
\newblock 2017.
\newblock Cross-lingual {RST} discourse parsing.
\newblock In {\em Proceedings of the 15th Conference of the {E}uropean Chapter
  of the Association for Computational Linguistics: Volume 1, Long Papers},
  pages 292--304, Valencia, Spain, April. Association for Computational
  Linguistics.

\bibitem[\protect\citename{Carlson \bgroup et al.\egroup }2002]{carlson2002rst}
Lynn Carlson, Mary~Ellen Okurowski, and Daniel Marcu.
\newblock 2002.
\newblock {\em {RST} discourse treebank}.
\newblock Linguistic Data Consortium, University of Pennsylvania.

\bibitem[\protect\citename{Chakrabarty \bgroup et al.\egroup
  }2019]{chakrabarty2019ampersand}
Tuhin Chakrabarty, Christopher Hidey, Smaranda Muresan, Kathleen McKeown, and
  Alyssa Hwang.
\newblock 2019.
\newblock Ampersand: Argument mining for persuasive online discussions.
\newblock In {\em Proceedings of the 2019 Conference on Empirical Methods in
  Natural Language Processing and the 9th International Joint Conference on
  Natural Language Processing (EMNLP-IJCNLP)}, pages 2926--2936.

\bibitem[\protect\citename{Devlin \bgroup et al.\egroup }2018]{devlin2018bert}
Jacob Devlin, Ming-Wei Chang, Kenton Lee, and Kristina Toutanova.
\newblock 2018.
\newblock Bert: Pre-training of deep bidirectional transformers for language
  understanding.
\newblock {\em arXiv preprint arXiv:1810.04805}.

\bibitem[\protect\citename{Diao \bgroup et al.\egroup }2014]{diao2014jointly}
Qiming Diao, Minghui Qiu, Chao-Yuan Wu, Alexander~J Smola, Jing Jiang, and
  Chong Wang.
\newblock 2014.
\newblock Jointly modeling aspects, ratings and sentiments for movie
  recommendation (jmars).
\newblock In {\em Proceedings of the 20th ACM SIGKDD international conference
  on Knowledge discovery and data mining}, pages 193--202. ACM.

\bibitem[\protect\citename{Feng and Hirst}2014]{feng-hirst-2014-linear}
Vanessa~Wei Feng and Graeme Hirst.
\newblock 2014.
\newblock A linear-time bottom-up discourse parser with constraints and
  post-editing.
\newblock In {\em Proceedings of the 52nd Annual Meeting of the Association for
  Computational Linguistics (Volume 1: Long Papers)}, pages 511--521,
  Baltimore, Maryland, June. Association for Computational Linguistics.

\bibitem[\protect\citename{Ferracane \bgroup et al.\egroup
  }2019]{ferracane2019evaluating}
Elisa Ferracane, Greg Durrett, Junyi~Jessy Li, and Katrin Erk.
\newblock 2019.
\newblock Evaluating discourse in structured text representations.
\newblock {\em arXiv preprint arXiv:1906.01472}.

\bibitem[\protect\citename{Gerani \bgroup et al.\egroup
  }2014]{gerani2014abstractive}
Shima Gerani, Yashar Mehdad, Giuseppe Carenini, Raymond~T Ng, and Bita Nejat.
\newblock 2014.
\newblock Abstractive summarization of product reviews using discourse
  structure.
\newblock In {\em Proceedings of the 2014 conference on empirical methods in
  natural language processing (EMNLP)}, pages 1602--1613.

\bibitem[\protect\citename{Hendrycks and Gimpel}2016]{hendrycks2016gaussian}
Dan Hendrycks and Kevin Gimpel.
\newblock 2016.
\newblock Gaussian error linear units (gelus).
\newblock {\em arXiv preprint arXiv:1606.08415}.

\bibitem[\protect\citename{Hernault \bgroup et al.\egroup
  }2010]{hernault2010hilda}
Hugo Hernault, Helmut Prendinger, Mitsuru Ishizuka, et~al.
\newblock 2010.
\newblock Hilda: A discourse parser using support vector machine
  classification.
\newblock {\em Dialogue \& Discourse}, 1(3).

\bibitem[\protect\citename{Hogenboom \bgroup et al.\egroup
  }2015]{hogenboom2015using}
Alexander Hogenboom, Flavius Frasincar, Franciska De~Jong, and Uzay Kaymak.
\newblock 2015.
\newblock Using rhetorical structure in sentiment analysis.
\newblock {\em Commun. ACM}, 58(7):69--77.

\bibitem[\protect\citename{Huber and Carenini}2020]{huber2020MEGA}
Patrick Huber and Giuseppe Carenini.
\newblock 2020.
\newblock Mega {RST} discourse treebanks with structure and nuclearity from
  scalable distant sentiment supervision.
\newblock In {\em Proceedings of the 2020 Conference on Empirical Methods in
  Natural Language Processing}.

\bibitem[\protect\citename{Ji and Eisenstein}2014]{ji2014representation}
Yangfeng Ji and Jacob Eisenstein.
\newblock 2014.
\newblock Representation learning for text-level discourse parsing.
\newblock In {\em Proceedings of the 52nd Annual Meeting of the Association for
  Computational Linguistics (Volume 1: Long Papers)}, volume~1, pages 13--24.

\bibitem[\protect\citename{Ji and Smith}2017]{ji2017neural}
Yangfeng Ji and Noah~A Smith.
\newblock 2017.
\newblock Neural discourse structure for text categorization.
\newblock In {\em Proceedings of the 55th Annual Meeting of the Association for
  Computational Linguistics (Volume 1: Long Papers)}, pages 996--1005.

\bibitem[\protect\citename{Joty \bgroup et al.\egroup }2015]{joty2015codra}
Shafiq Joty, Giuseppe Carenini, and Raymond~T Ng.
\newblock 2015.
\newblock Codra: A novel discriminative framework for rhetorical analysis.
\newblock {\em Computational Linguistics}, 41(3).

\bibitem[\protect\citename{Karimi and Tang}2019]{karimi2019learning}
Hamid Karimi and Jiliang Tang.
\newblock 2019.
\newblock Learning hierarchical discourse-level structure for fake news
  detection.
\newblock {\em arXiv preprint arXiv:1903.07389}.

\bibitem[\protect\citename{Li \bgroup et al.\egroup
  }2016]{li-etal-2016-discourse}
Qi~Li, Tianshi Li, and Baobao Chang.
\newblock 2016.
\newblock Discourse parsing with attention-based hierarchical neural networks.
\newblock In {\em Proceedings of the 2016 Conference on Empirical Methods in
  Natural Language Processing}, pages 362--371, Austin, Texas, November.
  Association for Computational Linguistics.

\bibitem[\protect\citename{Li \bgroup et al.\egroup
  }2018]{li-sun-joty-ijcai-18}
Jing Li, Aixin Sun, and Shafiq Joty.
\newblock 2018.
\newblock Segbot: A generic neural text segmentation model with pointer
  network.
\newblock In {\em Proceedings of the 27th International Joint Conference on
  Artificial Intelligence and the 23rd European Conference on Artificial
  Intelligence}, IJCAI-ECAI-2018, pages xx -- xx, Stockholm, Sweden, July.

\bibitem[\protect\citename{Lin \bgroup et al.\egroup }2019]{lin2019unified}
Xiang Lin, Shafiq Joty, Prathyusha Jwalapuram, and Saiful Bari.
\newblock 2019.
\newblock A unified linear-time framework for sentence-level discourse parsing.
\newblock {\em arXiv preprint arXiv:1905.05682}.

\bibitem[\protect\citename{Liu and Lapata}2018]{liu-lapata-2018-learning}
Yang Liu and Mirella Lapata.
\newblock 2018.
\newblock Learning structured text representations.
\newblock {\em Transactions of the Association for Computational Linguistics},
  6:63--75.

\bibitem[\protect\citename{Liu \bgroup et al.\egroup }2019a]{liu2019single}
Yang Liu, Ivan Titov, and Mirella Lapata.
\newblock 2019a.
\newblock Single document summarization as tree induction.
\newblock In {\em Proceedings of the 2019 Conference of the North American
  Chapter of the Association for Computational Linguistics: Human Language
  Technologies, Volume 1 (Long and Short Papers)}, pages 1745--1755.

\bibitem[\protect\citename{Liu \bgroup et al.\egroup }2019b]{liu2019roberta}
Yinhan Liu, Myle Ott, Naman Goyal, Jingfei Du, Mandar Joshi, Danqi Chen, Omer
  Levy, Mike Lewis, Luke Zettlemoyer, and Veselin Stoyanov.
\newblock 2019b.
\newblock Roberta: A robustly optimized bert pretraining approach.

\bibitem[\protect\citename{Loshchilov and Hutter}2019]{loshchilov2018decoupled}
Ilya Loshchilov and Frank Hutter.
\newblock 2019.
\newblock Decoupled weight decay regularization.
\newblock In {\em International Conference on Learning Representations}.

\bibitem[\protect\citename{Mabona \bgroup et al.\egroup
  }2019]{mabona-etal-2019-neural}
Amandla Mabona, Laura Rimell, Stephen Clark, and Andreas Vlachos.
\newblock 2019.
\newblock Neural generative rhetorical structure parsing.
\newblock In {\em Proceedings of the 2019 Conference on Empirical Methods in
  Natural Language Processing and the 9th International Joint Conference on
  Natural Language Processing (EMNLP-IJCNLP)}, pages 2284--2295, Hong Kong,
  China, November. Association for Computational Linguistics.

\bibitem[\protect\citename{Mann and Thompson}1988]{mann1988rhetorical}
William~C Mann and Sandra~A Thompson.
\newblock 1988.
\newblock Rhetorical structure theory: Toward a functional theory of text
  organization.
\newblock {\em Text-Interdisciplinary Journal for the Study of Discourse},
  8(3):243--281.

\bibitem[\protect\citename{Marcu}2000]{marcu-discourse}
Daniel Marcu.
\newblock 2000.
\newblock {\em The Theory and Practice of Discourse Parsing and Summarization}.
\newblock MIT Press, Cambridge, MA, USA.

\bibitem[\protect\citename{Morey \bgroup et al.\egroup
  }2017]{morey-etal-2017-much}
Mathieu Morey, Philippe Muller, and Nicholas Asher.
\newblock 2017.
\newblock How much progress have we made on {RST} discourse parsing? a
  replication study of recent results on the {RST-DT}.
\newblock In {\em Proceedings of the 2017 Conference on Empirical Methods in
  Natural Language Processing}, pages 1319--1324, Copenhagen, Denmark,
  September. Association for Computational Linguistics.

\bibitem[\protect\citename{Morey \bgroup et al.\egroup
  }2018]{morey2018dependency}
Mathieu Morey, Philippe Muller, and Nicholas Asher.
\newblock 2018.
\newblock A dependency perspective on {RST} discourse parsing and evaluation.
\newblock {\em Computational Linguistics}, 44(2):197--235.

\bibitem[\protect\citename{Nejat \bgroup et al.\egroup
  }2017]{nejat2017exploring}
Bita Nejat, Giuseppe Carenini, and Raymond Ng.
\newblock 2017.
\newblock Exploring joint neural model for sentence level discourse parsing and
  sentiment analysis.
\newblock In {\em Proceedings of the 18th Annual SIGdial Meeting on Discourse
  and Dialogue}, pages 289--298.

\bibitem[\protect\citename{Pascanu \bgroup et al.\egroup
  }2013]{pascanu2013difficulty}
Razvan Pascanu, Tomas Mikolov, and Yoshua Bengio.
\newblock 2013.
\newblock On the difficulty of training recurrent neural networks.
\newblock In {\em International conference on machine learning}, pages
  1310--1318.

\bibitem[\protect\citename{Paszke \bgroup et al.\egroup
  }2019]{NIPS2019_pytorch}
Adam Paszke, Sam Gross, Francisco Massa, Adam Lerer, James Bradbury, Gregory
  Chanan, Trevor Killeen, Zeming Lin, Natalia Gimelshein, Luca Antiga, Alban
  Desmaison, Andreas Kopf, Edward Yang, Zachary DeVito, Martin Raison, Alykhan
  Tejani, Sasank Chilamkurthy, Benoit Steiner, Lu~Fang, Junjie Bai, and Soumith
  Chintala.
\newblock 2019.
\newblock Pytorch: An imperative style, high-performance deep learning library.
\newblock In H.~Wallach, H.~Larochelle, A.~Beygelzimer, F.~d~Alch\'{e}-Buc,
  E.~Fox, and R.~Garnett, editors, {\em Advances in Neural Information
  Processing Systems 32}, pages 8026--8037. Curran Associates, Inc.

\bibitem[\protect\citename{Prasad \bgroup et al.\egroup }2008]{prasadpenn}
Rashmi Prasad, Nikhil Dinesh, Alan Lee, Eleni Miltsakaki, Livio Robaldo,
  Aravind Joshi, and Bonnie Webber.
\newblock 2008.
\newblock The penn discourse treebank 2.0.
\newblock {\em LREC}.

\bibitem[\protect\citename{Sanh \bgroup et al.\egroup
  }2019]{sanh2019distilbert}
Victor Sanh, Lysandre Debut, Julien Chaumond, and Thomas Wolf.
\newblock 2019.
\newblock Distilbert, a distilled version of bert: smaller, faster, cheaper and
  lighter.

\bibitem[\protect\citename{Subba and Di~Eugenio}2009]{subba2009effective}
Rajen Subba and Barbara Di~Eugenio.
\newblock 2009.
\newblock An effective discourse parser that uses rich linguistic information.
\newblock In {\em Proceedings of Human Language Technologies: The 2009 Annual
  Conference of the North American Chapter of the Association for Computational
  Linguistics}, pages 566--574. Association for Computational Linguistics.

\bibitem[\protect\citename{Tang \bgroup et al.\egroup }2015]{tang2015document}
Duyu Tang, Bing Qin, and Ting Liu.
\newblock 2015.
\newblock Document modeling with gated recurrent neural network for sentiment
  classification.
\newblock In {\em Proceedings of the 2015 conference on empirical methods in
  natural language processing}, pages 1422--1432.

\bibitem[\protect\citename{Vaswani \bgroup et al.\egroup }2017]{NIPS2017_7181}
Ashish Vaswani, Noam Shazeer, Niki Parmar, Jakob Uszkoreit, Llion Jones,
  Aidan~N Gomez, \L~ukasz Kaiser, and Illia Polosukhin.
\newblock 2017.
\newblock Attention is all you need.
\newblock In I.~Guyon, U.~V. Luxburg, S.~Bengio, H.~Wallach, R.~Fergus,
  S.~Vishwanathan, and R.~Garnett, editors, {\em Advances in Neural Information
  Processing Systems 30}, pages 5998--6008. Curran Associates, Inc.

\bibitem[\protect\citename{Wang \bgroup et al.\egroup
  }2017]{wang-etal-2017-two}
Yizhong Wang, Sujian Li, and Houfeng Wang.
\newblock 2017.
\newblock A two-stage parsing method for text-level discourse analysis.
\newblock In {\em Proceedings of the 55th Annual Meeting of the Association for
  Computational Linguistics (Volume 2: Short Papers)}, pages 184--188,
  Vancouver, Canada, July. Association for Computational Linguistics.

\bibitem[\protect\citename{Wolf \bgroup et al.\egroup
  }2019]{Wolf2019HuggingFacesTS}
Thomas Wolf, Lysandre Debut, Victor Sanh, Julien Chaumond, Clement Delangue,
  Anthony Moi, Pierric Cistac, Tim Rault, R'emi Louf, Morgan Funtowicz, and
  Jamie Brew.
\newblock 2019.
\newblock Huggingface's transformers: State-of-the-art natural language
  processing.
\newblock {\em ArXiv}, abs/1910.03771.

\bibitem[\protect\citename{Yu \bgroup et al.\egroup }2018]{yu2018transition}
Nan Yu, Meishan Zhang, and Guohong Fu.
\newblock 2018.
\newblock Transition-based neural {RST} parsing with implicit syntax features.
\newblock In {\em Proceedings of the 27th International Conference on
  Computational Linguistics}, pages 559--570.

\bibitem[\protect\citename{Zhang \bgroup et al.\egroup
  }2015]{zhang2015character}
Xiang Zhang, Junbo Zhao, and Yann LeCun.
\newblock 2015.
\newblock Character-level convolutional networks for text classification.
\newblock In {\em Advances in neural information processing systems}, pages
  649--657.

\end{thebibliography}

\end{document}